\documentclass[runningheads]{llncs}
\usepackage{algorithm}
\usepackage{algorithmicx}
\usepackage{algpseudocode}
\usepackage{amsmath}
\floatname{algorithm}{Algorithm}

\usepackage{graphicx}
\usepackage{multirow}
\usepackage{tabularx}
\usepackage{arydshln}
\begin{document}
%
\title{Abstractive Summarization Improved by WordNet-based Extractive Sentences}
\titlerunning{Abstractive summarization with extractive methods}
%
\author{Niantao Xie\inst{1} \and
Sujian Li\inst{1} \and
Huiling Ren\inst{2} \and
Qibin Zhai\inst{3}}
\authorrunning{Niantao Xie et al.}
%
\institute{MOE Key Laboratory of Computational Linguistics, Peking University, China \and
Institute of Medical Information, Chinese Academy of Medical Sciences \and
MOE Information Security Lab, School of Software \& Microelectronics, Peking University, China\\
\email{\{xieniantao,lisujian\}@pku.edu.cn}\\
\email{ren.huiling@imicams.ac.cn}\\
\email{qibinzhai@ss.pku.edu.cn}}
\maketitle              
\begin{abstract}
Recently, the \emph{seq2seq} abstractive summarization models have achieved good results on the CNN/Daily Mail dataset.
Still, how to improve abstractive methods with extractive methods is a good research direction, since extractive methods have their potentials of exploiting various efficient features for extracting important sentences in one text. 
In this paper, in order to improve the semantic relevance of abstractive summaries, we adopt the \emph{WordNet\ based\ sentence\ ranking\ algorithm} to extract the sentences which are most semantically to one text. Then, we design a \emph{dual\ attentional\ seq2seq\ framework} to generate summaries with consideration of the extracted information. 
At the same time, we combine \emph{pointer-generator} and \emph{coverage} mechanisms to solve the problems of out-of-vocabulary (OOV) words and duplicate words which exist in the abstractive models. Experiments on the CNN/Daily Mail dataset show that our models achieve competitive performance with the state-of-the-art ROUGE scores. Human evaluations also show that the summaries generated by our models have high semantic relevance to the original text.


\end{abstract}
\section{Introduction}
For automatic summarization, there are two main methods: extractive and abstractive.
Extractive methods use certain scoring rules or ranking methods to select a certain number of important sentences from the source texts. For example, \cite{cao2016attsum} proposed to make use of Convolutional Neural Networks (CNN) to represent queries and sentences, as well as adopted a greedy algorithm combined with \emph{pair-wise ranking algorithm} for extraction. Based on Recurrent Neural Networks (RNN), \cite{nallapati2017summarunner} constructed a sequence classifier and obtained the highest extractive scores on the CNN/Daily Mail corpus set. At the same time, The abstractive summarization models attempt to simulate the process of how human beings write summaries and need to analyze, paraphrase, and reorganize the source texts. It is known that there exist two main problems called OOV words and duplicate words by means of abstraction. \cite{see2017get} proposed an improved pointer mechanism named \emph{pointer-generator} to solve the OOV words as well as came up with a variant of coverage vector called \emph{coverage} to deal with the duplicate words. \cite{nema2017diversity} created the \emph{diverse cell} structures to handle duplicate words problem based on query-based summarization. For the first time,  a \emph{reinforcement learning} method based neural network model was raised and obtained the state-of-the-art scores on the CNN/Daily Mail corpus\cite{paulus2017deep}.

Both extractive and abstractive methods have their merits. In this paper, we employ the combination of extractive and abstractive methods at the sentence level. In the extractive process, we find that there are some ambiguous words in the source texts. The different meanings of each word can be acquired through the synonym dictionary called WordNet. First \emph{WordNet based Lesk algorithm} is utilized to analyze the word semantics. Then we apply the \emph{modified sentence ranking algorithm} to extract a specified number of sentences according to the sentence syntactic information. During the abstractive part based on \emph{seq2seq model}, we add a new encoder which is derived from the extractive sentences and put the \emph{dual attention} mechanism  for decoding operations. As far as we know, it is the first time that joint training of sentence-level extractive and abstractive models has been conducted. Additionally, we combine the \emph{pointer-generator} and \emph{coverage} mechanisms to handle the OOV words and duplicate words.

Our contributions in this paper are mainly summarized as follows:
\begin{itemize}
\item Considering the semantics of words and sentences, we improve the \emph{sentence ranking algorithm} based on the \emph{WordNet-based simplified lesk algorithm} to obtain important sentences from the source texts.
\item We construct two parallel encoders from the extracted sentences and source texts separately, and make use of \emph{seq2seq dual attentional model} for joint training.
\item We adopt the \emph{pointer-generator} and \emph{coverage} mechanisms to deal with OOV words and duplicate words problems. Our results are competitive compared with the state-of-the-art scores.
\end{itemize}

\section{Our Method}
\begin{figure}[h!]
\begin{center}
\includegraphics[width=0.95\textwidth]{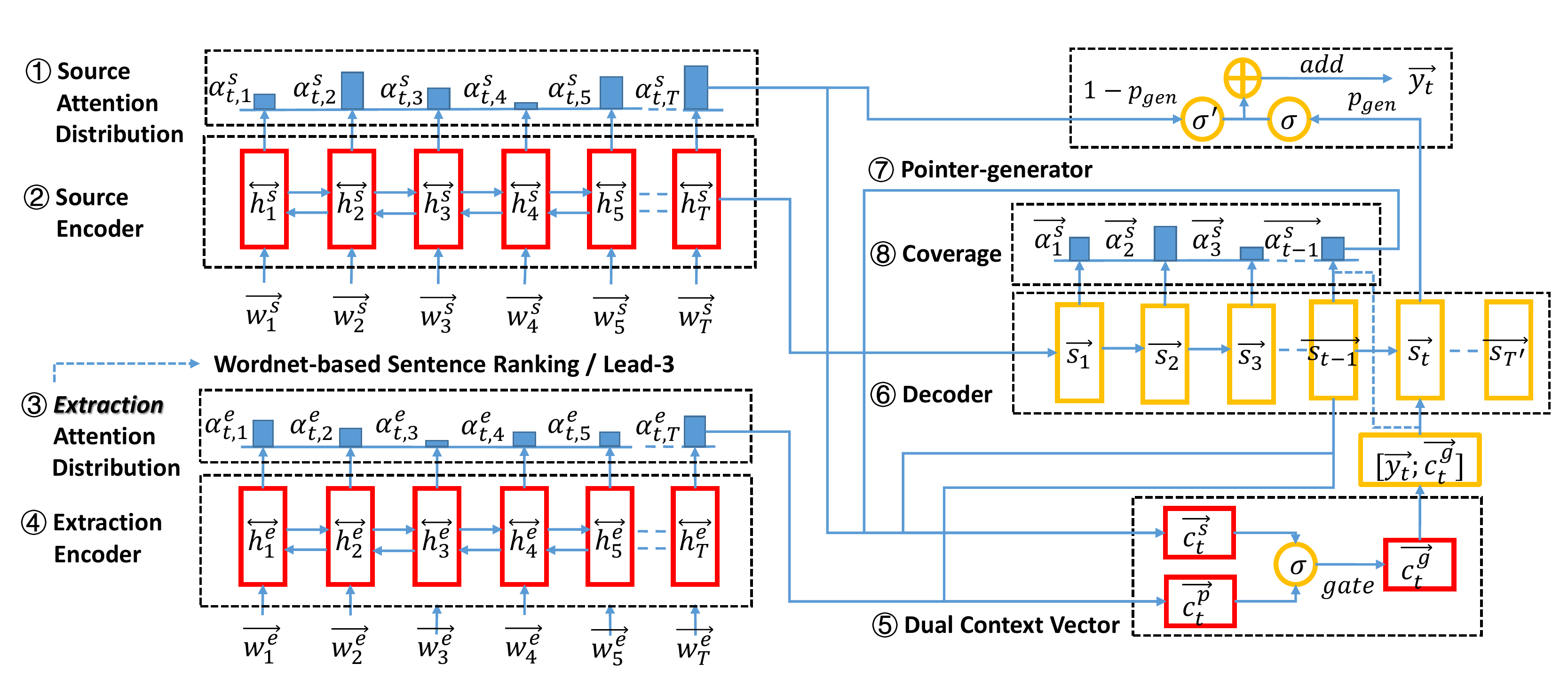}
\caption{A \emph{dual attentional encoders-decoder model} with \emph{pointer-generator} network.} \label{fig1}
\end{center}
\end{figure}

Our method is based on the \emph{seq2seq attentional model}, which is implemented with reference to \cite{nallapati2016abstractive} and the attention distribution $\vec{\alpha}_t$ is calculated as in \cite{bahdanau2014neural}. Here, we show the architecture of our model which is composed of eight parts as in Figure \ref{fig1}. We construct two encoders (\textcircled{\scriptsize 2}\textcircled{\scriptsize 4}) based on the source texts and extracted sentences, as well as take advantage of a \emph{dual attentional} decoder (\textcircled{\scriptsize 1} \textcircled{\scriptsize 3}\textcircled{\scriptsize 5}\textcircled{\scriptsize 6}) to generate summaries. Finally, we combine the \emph{pointer-generator} (\textcircled{\scriptsize 7}) and \emph{coverage} mechanisms (\textcircled{\scriptsize 8}) to manage OOV and duplicate words problems.

\subsection{Seq2seq dual attentional model}
\subsubsection{Encoders-decoder model. }
Referring to \cite{bahdanau2014neural}, we use two single-layer bidirectional Long Short-Term Memory (BiLSTM) encoders including source and extractive encoders, and a single-layer unidirectional LSTM (UniLSTM) decoder in our model, as shown in Figure \ref{fig1}. For encoding time $i$, the source texts and the extracted information respectively input the word embeddings $\vec{w}_i^s$ and $\vec{w}_i^e$ into two encoders. Meanwhile, the corresponding hidden layer states $\overleftrightarrow{\vec{h}}_i^s$ and $\overleftrightarrow{\vec{h}}_i^e$ are generated. At decoding step $t$, the decoder will receive the word embedding from the step $t-1$, which is obtained according to the previous word in the reference summary during training, or provided by the decoder itself when testing. Next we acquire the state $\vec{s}_t$ and produce the vocabulary distribution $P({\vec{y}_t})$.

Here, we are supposed to calculate $\overleftrightarrow{\vec{h}}_i^s$ by the following formulas:
\begin{equation}
\overrightarrow{\vec{h}}_i^s = LSTM(\vec{w}_i^s,
\overrightarrow{\vec{h}}_{i-1}^s)
\end{equation}
\begin{equation}
\overleftarrow{\vec{h}}_i^s = LSTM(\vec{w}_i^s, \overleftarrow{\vec{h}}_{i+1}^s)
\end{equation}
\begin{equation}
\overleftrightarrow{\vec{h}}_i^s = [\overrightarrow{\vec{h}}_i^s;\  \overleftarrow{\vec{h}}_i^s]
\end{equation}

Also, $\overleftrightarrow{\vec{h}}_i^e$ could be obtained as follows:
\begin{equation}
\overrightarrow{\vec{h}}_i^e = LSTM(\vec{w}_i^e,
\overrightarrow{\vec{h}}_{i-1}^e)
\end{equation}
\begin{equation}
\overleftarrow{\vec{h}}_i^e = LSTM(\vec{w}_i^e, \overleftarrow{\vec{h}}_{i+1}^e)
\end{equation}
\begin{equation}
\overleftrightarrow{\vec{h}}_i^e = [\overrightarrow{\vec{h}}_i^e;\  \overleftarrow{\vec{h}}_i^e]
\end{equation}

\subsubsection{Dual attention mechanism. }
At the $t^{th}$ step, we need not only the previous hidden state $\vec{s}_{t-1}$, but also the context vector $\vec{c}_{t-1}^s$, $\vec{c}_{t-1}^e$, $\vec{c}_t^s$, $\vec{c}_t^e$ obtained by the corresponding attention distribution \cite{bahdanau2014neural} to gain state $\vec{s}_t$ and vocabulary distribution $P({\vec{y}_t})$.

Firstly, for source encoder, we calculate the context vector ${\vec{c}}_t^s$ in the following way ($\textbf{V}^s$, $\textbf{W}_1^s$, $\textbf{W}_2^s$, $\vec{b}^s$ are learnable parameters):
\begin{equation}
{{e}}_{i,\ t}^s = {\textbf{V}^s}^T\cdot tanh{(\textbf{W}_1^s\cdot\vec{s}_t+\textbf{W}_2^s\cdot\overleftrightarrow{\vec{h}}_i^s+\vec{b}^s)}
\end{equation}
\begin{equation}
{{\alpha}}_{i,\ t}^s = \frac{{{e}}_{i,\ t}^s}{\sum_{j=1}^{n_s}{{e}}_{j,\ t}^s}
\end{equation}
\begin{equation}
{\vec{c}}_t^s = \sum_{i=1}^{n_s}{{\alpha}}_{i,\ t}^s\cdot\overleftrightarrow{\vec{h}}_t^s
\end{equation}

Secondly, for extractive encoder, we utilize the identical method to compute the context vector ${\vec{c}}_t^e$ ($\textbf{V}^e$, $\textbf{W}_1^e$, $\textbf{W}_2^e$, $\vec{b}^e$ are learnable parameters):
\begin{equation}
{{e}}_{i,\ t}^e = {\textbf{V}^e}^T\cdot tanh{(\textbf{W}_1^e\cdot\vec{s}_t+\textbf{W}_2^e\cdot\overleftrightarrow{\vec{h}}_i^e+\vec{b}^e)}
\end{equation}
\begin{equation}
{{\alpha}}_{i,\ t}^e = \frac{{{e}}_{i,\ t}^e}{\sum_{j=1}^{n_e}{{e}}_{j,\ t}^e}
\end{equation}
\begin{equation}
{\vec{c}}_t^e = \sum_{i=1}^{n_e}{{\alpha}}_{i,\ t}^e\cdot\overleftrightarrow{\vec{h}}_t^e
\end{equation}

Thirdly, we get the gated context vector ${\vec{c}}_t^g$ by calculating the weighted sum of context vectors ${\vec{c}}_t^s$ and ${\vec{c}}_t^e$, where the weight is the \emph{gate network} obtained by the concatenation of ${\vec{c}}_t^s$ and ${\vec{c}}_t^e$ via \emph{multi-layer perceptron (MLP)}. Details are shown as below  ($\sigma$ is Sigmoid function, $\textbf{W}^g$, $\vec{b}^g$ are learnable parameters):
\begin{equation}
\vec{g}_t = \sigma(\textbf{W}^g\cdot[\vec{c}_t^s;\ \vec{c}_t^e]+\vec{b}^g)
\end{equation}
\begin{equation}
{\vec{c}}_t^{g} = \vec{g}_t\cdot{\vec{c}}_t^s+(\vec{1}-\vec{g}_t)\cdot{\vec{c}}_t^e
\end{equation}

In the same way, we can obtain the hidden state ${\vec{s}}_t$ and predicte the probability distribution $P({\vec{y}_t})$ at time $t$  ($\textbf{W}_1^{in}$, $\textbf{W}_2^{in}$, $\vec{b}_{in}$, $\textbf{W}_1^{out}$, $\textbf{W}_2^{out}$, $\vec{b}_{out}$ are learnable parameters).
\begin{equation}
{\vec{s}}_t = LSTM({\vec{s}}_{t-1},\ \textbf{W}_1^{in}\cdot{\vec{y}}_{t-1}+\textbf{W}_2^{in}\cdot{\vec{c}}_{t-1}^{g}+\vec{b}_{in})
\end{equation}
\begin{equation}
P({\vec{y}}_t | {\vec{y}}_{<t}, \vec{x}) = softmax(\textbf{W}_1^{out}\cdot{\vec{s}}_t+\textbf{W}_2^{out}\cdot{\vec{c}}_t^{g}+\vec{b}_{out})
\end{equation}

\subsection{WordNet-based Sentence Ranking Algorithm}
To extract the important sentences, we adopt a \emph{WordNet-based sentence ranking algorithm}.
WordNet\footnote{\tt http://www.nltk.org/howto/wordnet.html} is a lexical database for the English language, which groups English words into sets of synonyms called synsets and provides short definitions and usage examples. \cite{pal2014approach} used the \emph{simplified lesk approach} based on WordNet to extract abstracts. We refer to its algorithm and set up our \emph{sentence ranking algorithm} so as to construct the extractive encoder.

For sentence $\vec{x} = (x_1, x_2, ..., x_n)$, after filtering out the stop words and unambiguous tokens through WordNet, we obtain a reserved subsequence $\vec{x}^{'} = (x_{i_1}, x_{i_2}, ..., x_{i_m})$. 
Since some words contain too many different senses which may result in too much calculation, 
we set a window size $n_{win}$ (default value is 5) and sort $\vec{x}^{'}$ in descending order according to the number of senses of words, as well as keep the first $n_{sav}$ ($n_{sav} = min(m, n_{win})$) words left to get $\vec{x}^{''} = (x_{s_1}, x_{s_2}, ..., x_{s_{n_{sav}}})$. Next, we count the common number of senses of each word as word weight. Finally, we get the sum weights of each sentence and acquire an average sentence weight.

Taking a sentence $\vec{x}^{''} = (x_1, x_2, x_3)$ for instance, we make an assumption that $x_1$ has two senses $\vec{m}_a$ and $\vec{m}_b$, $x_2$ has two senses $\vec{m}_c$ and $\vec{m}_d$, while $x_3$ has two senses $\vec{m}_e$, $\vec{m}_f$. Currently considering $x_1$ as the keyword, we measure the number of common words between a pair of sentences, which describe the word senses of $x_1$ and another word.

Table \ref{tab1} shows all possible matches of the senses of $x_1$, $x_2$, $x_3$. For the two senses of $x_1$, we can separately obtain the sum of co-occurrence word pairs for each meaning. For $\vec{m}_a$, we obtain $count_{\vec{m}_a}$ = $count_{ac}$ + $count_{ad}$ + $count_{ae}$ + $count_{af}$, for $\vec{m}_b$, we gain $count_{\vec{m}_b}$ = $count_{bc}$ + $count_{bd}$ + $count_{be}$ + $count_{bf}$. The significance corresponding to the higher score $count_{x_1} $( $count_{\vec{m}_a}$ or $count_{\vec{m}_b}$) is assigned to the the keyword $x_1$.

\begin{table}[h!]
\begin{center}
\caption{The number of common words between a pair of sentences.}\label{tab1}
\begin{tabular}{c|c}
\hline
\ Pair of sentences \  &  \ common words in sense description \ \\
\hline
$\vec{m}_a$ and $\vec{m}_c$ & $count_{ac}$\\
$\vec{m}_a$ and $\vec{m}_d$ & $count_{ad}$\\
$\vec{m}_b$ and $\vec{m}_c$ & $count_{bc}$\\
$\vec{m}_b$ and $\vec{m}_d$ & $count_{bd}$\\
$\vec{m}_a$ and $\vec{m}_e$ & $count_{ae}$\\
$\vec{m}_a$ and $\vec{m}_f$ & $count_{af}$\\
$\vec{m}_b$ and $\vec{m}_e$ & $count_{be}$\\
$\vec{m}_b$ and $\vec{m}_f$ & $count_{bf}$\\
\hline
\end{tabular}
\end{center}
\end{table}

In this way, we're capable of acquiring the average weight of sentence $\vec{x}$.
\begin{equation}
weight_{avg} = \frac{1}{n_{sav}}\sum_{i=1}^{n_{sav}} count_{x_i}
\end{equation}

Let's assume that document $\vec{D} = (\vec{x}_1, \vec{x}_2, ..., \vec{x}_N)$, which contains a total of $N$ sentences. We sort them in descending order according to the average weights of sentences, and then extract the top $n_{top}$ sentences (default value is 3).

\subsection{Pointer-generator and coverage mechanisms}
\subsubsection{Pointer-generator network. }
\emph{Pointer-generator} is an effective method to solve the problem of OOV words and its structure has been expanded in Figure \ref{fig1}. We borrow the method improved by \cite{see2017get}. $p_{gen}$ is defined as a switch to decide to generate a word from the vocabulary or copy a word from the source encoder attention distribution. We maintain an extended vocabulary including the vocabulary and all words in the source texts. For the decoding step $t$ and decoder input $\vec{x}_t$, we define $p_{gen}$ as:
\begin{equation}
p_{gen} = \sigma(\textbf{W}_1^p\cdot\vec{c}_t^s+\textbf{W}_2^p\cdot\vec{s}_t+\textbf{W}_3^p\cdot\vec{x}_t+\vec{b}^p)
\end{equation}
\begin{equation}
P_{vocab} = P({\vec{y}}_t | {\vec{y}}_{<t}, \vec{x})
\end{equation}
\begin{equation}
P(\vec{w}_t) = p_{gen}P_{vocab}(\vec{w}_t)+(1-p_{gen})\sum_{i: \vec{w}_i=\vec{w}_t}{\alpha}_{i,\ t}^s
\end{equation}

Where $\vec{w}_t$ is the value of $\vec{x}_t$, and $\textbf{W}_1^p$, $\textbf{W}_2^p$, $\textbf{W}_3^p$, $\vec{b}^p$ are learnable parameters.

\subsubsection{Coverage mechanism. }
Duplicate words are a critical problem in the \emph{seq2seq model}, and even more serious when generating long texts like multi-sentence texts. \cite{see2017get} made some minor modifications to the \emph{coverage model} \cite{tu2016modeling} which is also displayed in Figure \ref{fig1}. 

First, we calculate the sum of attention distributions from previous decoder steps ($1, 2, 3, ..., t-1$) to get  a coverage vector $\vec{cov}_t$:
\begin{equation}
\vec{cov}_t^s = \sum_{t'=0}^{t-1}\vec{\alpha}_{t'}^s
\end{equation}

Then, we make use of coverage vector $\vec{cov}_t$ to update the attention distribution:
\begin{equation}
{{e}}_{i,\ t}^s = {\textbf{V}^s}^T\cdot tanh{(\textbf{W}_1^s\cdot\vec{s}_t+\textbf{W}_2^s\cdot\overleftrightarrow{\vec{h}}_i^s+\textbf{W}_3^s\cdot{cov}_{i,\ t}^s+\vec{b}^s)}
\end{equation}

Finally, we define the coverage loss function $covloss_t$ for the sake of penalizing the duplicate words appearing at decoding time $t$, and renew the total loss:
\begin{equation}
covloss_t = \sum_{i}min({\alpha}_{i,\ t}^s,\ {cov}_{i,\ t}^s)
\end{equation}
\begin{equation}
loss_t = -log(P(\vec{w}_t^*)) + {\lambda}\ covloss_t
\end{equation}

Where $\vec{w}_t^*$ is the target word at $t^{th}$ step, $-log(P(\vec{w}_t^*))$ is the primary loss for timestep $t$ during training, hyperparameter $\lambda$ (default value is 1.0) is the weight for $covloss_t$, $\textbf{W}_1^s$, $\textbf{W}_2^s$, $\textbf{W}_3^s$, $\vec{b}^s$ are learnable parameters.

\section{Experiments}
\subsection{Dataset}
CNN/Daily Mail dataset\footnote{\tt https://cs.nyu.edu/\~{}kcho/DMQA/} is widely used in the public automatic summarization evaluation, which contains online news articles (781 tokens on average) paired with multi-sentence summaries (56 tokens on average). \cite{see2017get} provided the data processing script, and we take advantage of it to obtain the non-anonymized version of the the data including 287,226 training pairs, 13,368 validation pairs and 11,490 test pairs, though \cite{nallapati2017summarunner,nallapati2016abstractive} used the anonymized version. During training steps, we find that 114 of 287,226 articles are empty, so we utilize the remaining 287,112 pairs for training. Then, we perform the splitting preprocessing for the data pairs with the help of Stanford CoreNLP toolkit\footnote{\tt https://stanfordnlp.github.io/CoreNLP/}, and	convert them into binary files, as well as get the vocab file for the convenience of reading data.

\subsection{Implementation}
\subsubsection{Model parameters configuration. }
The corresponding parameters of controlled experimental models are described as follows. For all models, we have set the word embeddings and RNN hidden states to be 128-dimensional and 256-dimensional respectively for source encoders, extractive encoders and decoders. Contrary to \cite{nallapati2016abstractive}, we learn the word embeddings from scratch during training, because our training dataset is large enough. We apply the optimization technique Adagrad with learning rate 0.15 and an initial accumulator value of 0.1, as well as employ the gradient clipping with a maximum gradient norm of 2.

For the one-encoder models, we set up the vocabulary size to be 50k for source encoder and target decoder simultaneously. We try to adjust the vocabulary size to be 150k, then discover that when the model is trained to converge, the time cost is doubled but the test dataset scores have slightly dropped. In our analysis, the models' parameters have increased excessively when the vocabulary enlarges, leading to overfitting during the training process. Meanwhile, for the models with two encoders, we adjust the vocabulary size to be 40k.

Each pair of the dataset consists of an article and a multi-sentence summary. We truncate the article to 400 tokens and limit the summary to 100 tokens for both training and testing time. During decoding mode, we generate at least 35 words with \emph{beam search algorithm}. Data truncation operations not only reduce memory consumption, speed up training and testing, but also improve the experimental results. The reason is that the vital information of news texts is mainly concentrated in the first half part.

We train on a single GeForce GTX 1080 GPU with a memory of 8114 MiB, and the batch size is set to be 16, as well as the beam size is 4 for \emph{beam search} in decoding mode. For the \emph{seq2seq dual attentional models} without \emph{pointer-generator}, we trained them for about two days. Models with \emph{pointer-generator} expedite the training, the time cost is reduced to about one day. When we add \emph{coverage}, the coverage loss weight $\lambda$ is set to 1.0, and the model needs about one hour for training.

\subsubsection{Controlled experiments. }
In order to figure out how each part of our models contributes to the test results, based on the released codes\footnote{\tt https://github.com/tensorflow/models/tree/master/research/textsum} of Tensorflow, we have implemented all the models and done a series of experiments.

The baseline model is a general \emph{seq2seq attentional model}, the encoder consists of a biLSTM and the decoder is made up of an uniLSTM. The second baseline model is our \emph{encoders-decoder dual attention model}, which contains two biLSTM encoders and one uniLSTM decoder. This model combines the extractive and generative methods to perform joint training effectively through a \emph{dual attention mechanism}. 

For the above two basic models, in order to explain how the OOV and duplicate words are treated, we lead into the \emph{pointer-generator} and \emph{coverage} mechanism step by step. For the second baseline, the two tricks are only related to the source encoder, because we think that the source encoder already covers all the tokens in the extractive encoder. For the extractive encoder, we adopt two methods for extraction. One is the \emph{leading three (lead-3)} sentences technique, which is simple but indeed a strong baseline. The other is the \emph{Modified sentence ranking algorithm} based on WordNet that we explain in details in section 3. It considers semantic relations in words and sentences from source texts.

\subsection{Results}
ROUGE \cite{lin2004rouge} is a set of metrics with a software package used for evaluating automatic summarization and machine translation results. It counts the number of overlapping basic units including n-grams, longest common subsequences (LCS). We use pyrouge\footnote{\tt https://pypi.org/project/pyrouge/0.1.3/}, a python wrapper to gain ROUGE-1, ROUGE-2 and ROUGE-L scores and list the $\textbf{F}_1$ scores in table \ref{tab2}.

\begin{table}[h!]
\begin{center}
\caption{ROUGE $\textbf{F}_1$ scores on CNN/Daily Mail non-anonymized testing dataset for all the controlled experiment models mentioned above. According to the official ROUGE usage description, all our ROUGE scores have a 95\% confidence interval of at most $\pm$0.25. \emph{PGN}, \emph{Cov}, \emph{ML}, \emph{RL} are abbreviations for \emph{pointer-generator}, \emph{coverage}, \emph{mixed-objective learning} and \emph{reinforcement learning}. Models with subscript $_a$ were trained and tested on the anonymized CNN/Daily Mail dataset, as well as with $^*$ are the state-of-the-art extractive and abstractive summarization models on the anonymized dataset by now. } \label{tab2}
\begin{tabular*}{\textwidth}{p{0.50\textwidth}|>{\hfil}p{0.167\textwidth}<{\hfil}|>{\hfil}p{0.167\textwidth}<{\hfil}|>{\hfil}p{0.130\textwidth}<{\hfil}}
\hline
\multirow{2}{*}{\ \ Models}    & \multicolumn{3}{c}{ROUGE $\textbf{F}_1$ scores} \\
\cline{2-4} & 1 & 2 & L \\
\hline
\ \ Seq2seq + Attn & 31.50 & 11.95 & 28.85 \\
\ \ Seq2seq + Attn (150k) & 30.67 & 11.32 & 28.11 \\
\ \ Seq2seq + Attn + PGN & 36.58 & 15.76 & 33.33 \\
\ \ Seq2seq + Attn + PGN + Cov & \textbf{39.16} & \textbf{16.98} & \textbf{35.81} \\
\hline
\ \ Lead-3 + Dual-attn + PGN & 37.26 & 16.12 & 33.87 \\
\ \ WordNet + Dual-attn + PGN & 36.91 & 15.97 & 33.58 \\
\ \ Lead-3 + Dual-attn + PGN + Cov & \textbf{39.41} & \textbf{17.30} & 35.92 \\
\ \ WordNet + Dual-attn + PGN + Cov & 39.32 & 17.15 & \textbf{36.02} \\
\hline
\ \ Lead-3 (\cite{see2017get}) & 40.34 & 17.70 & 36.57 \\
\ \ Lead-3 (\cite{nallapati2017summarunner})$_{a}$ & 39.20 & 15.70 & 35.50 \\
\ \ SummaRuNNer (\cite{nallapati2017summarunner})$_{a}^*$ & \textbf{39.60} & \textbf{16.20} & \textbf{35.30} \\
\cdashline{1-4}[2pt/2pt]
\ \ RL + Intra-attn (\cite{paulus2017deep})$_{a}^*$ & \textbf{41.16} & 15.75 & \textbf{39.08} \\
\ \ ML + RL + Intra-attn (\cite{paulus2017deep})$_{a}$ & 39.87 & \textbf{15.82} & 36.90 \\
\hline
\end{tabular*}
\end{center}
\end{table}

We carry out the experiments based on original dataset, i.e., non-anonymized version of data. For the top three models in table \ref{tab2}, their ROUGE scores are slightly higher than those executed by \cite{see2017get}, except for the ROUGE-L score of \emph{Seq2seq + Attn + PGN}, which is 0.09 points lower than the former result. For the fourth model, we did not reproduce the results of \cite{see2017get}, ROUGE-1, ROUGE-2, and ROUGE-L decreased by an average of 0.41 points.

For the four models in the middle, we apply the \emph{dual attention} mechanism to integrate extraction with abstraction for joint training and decoding. These model variants own a single \emph{PGN} or \emph{PGN} together with \emph{Cov}, achieve better results than the corresponding vulgaris attentional models simultaneously. We conclude that the extractive encoders play a role, among which we obtained higher ROUGE-1 and ROUGE-2 scores based on the \emph{Lead-3 + Dual-attn + PGN + Cov} model, and achieve a better ROUGE-L score on \emph{WordNet + Dual-attn + PGN + Cov} model.

Let's take a look at the five models at the bottom, two of which give the state-of-the-art scores for the extractive and generative methods. our scores are already comparable to them. It is worthy to mention that based on the \emph{dual attention}, our models related to both \emph{Lead-3} and \emph{WordNet} with \emph{PGN} and \emph{Cov} have exceeded the previous best ROUGE-2 scores. When in fact, previous \emph{SummaRuNNer}, \emph{RL} related models are based on anonymized dataset, these differences may cause some deviations in the comparison of experimental results.

\begin{figure}[h!]
\begin{center}
\includegraphics[width=0.93\textwidth]{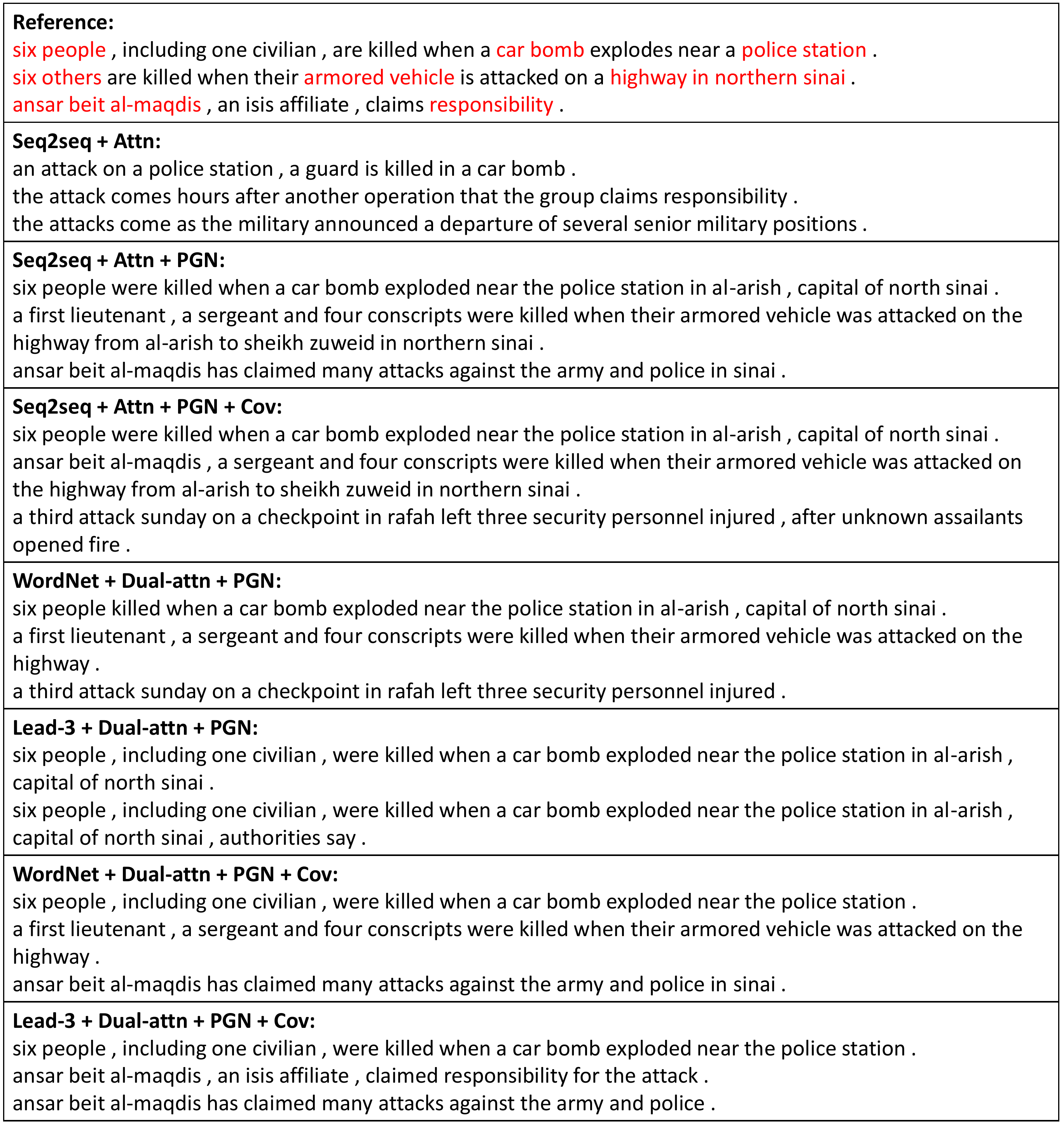}
\caption{Summaries for all the models of one test article example.} \label{fig2}
\end{center}
\end{figure}

We give some generated summaries of different models for one selected test article. From Figure \ref{fig2}, we can see that the red words represent key information about \emph{who}, \emph{what}, \emph{where} and \emph{when}. We can match the corresponding keywords in the remaining seven summaries to find out whether they cover all the significant points, and check if they are expressed in a concise and coherent way. It can be discovered from Figure \ref{fig2} that most of the models have lost several vital points, and the model \emph{Lead-3 + Dual-attn +  PGN} has undergone fairly serious repetition. Our model \emph{WordNet + Dual-attn + PGN + Cov} holds the main key information as well as has better readability and semantic correctness reliably.

\section{Related Work}
Up to now, automatic summarization with extractive and abstractive methods are under fervent research. On the one hand, the extractive techniques extract the topic-related keywords and significant sentences from the source texts to constitute summaries. \cite{cheng2016neural} proposed a \emph{seq2seq model} with a hierarchical encoder and attentional decoder to solve extractive summarization tasks at the word and sentence levels. Currently \cite{nallapati2017summarunner} put forward \emph{SummaRuNNer}, a RNN based sequence model for extractive summarization and it achieves the previous state-of-the-art performance. On the other hand, abstractive methods establish an intrinsic semantic representation and use natural language generation techniques to produce summaries which are closer to what human beings express. \cite{bahdanau2014neural} applied the combination of \emph{seq2seq model} and \emph{attention} mechanism to machine translation tasks for the first time. \cite{rush2015neural} exploited \emph{seq2seq model} to sentence compression to lay the groundwork for subsequent summarization with different granularities. \cite{lopyrev2015generating} used \emph{encoder-decoder with attention} method to generate news headlines. \cite{zhou2017selective} added a \emph{selective gate network} to the basic model in order to control which part of the information flowed from encoder to decoder. \cite{tan2017abstractive} raised a model based on graph and \emph{attention} mechanism to strengthen the positioning of vital information of source texts.

So as to solve rare and unseen words, \cite{gu2016incorporating,gulcehre2016pointing} proposed the \emph{COPYNET model} and \emph{pointing} mechanism, \cite{zeng2016efficient} created read-again and copy mechanisms. \cite{nallapati2016abstractive} made a combination of the basic model with \emph{large vocabulary trick (LVT)}, \emph{feature-rich encoder}, \emph{pointer-generator}, and \emph{hierarchical attention}. In addition to \emph{pointer-generator}, other tricks of this paper also contributed to the experiment results. \cite{see2017get} presented an updated version of \emph{pointer-generator} which proved to be better. As for duplicate words, for sake of solving problems of over or missing translation, \cite{tu2016modeling} came up with a \emph{coverage} mechanism to avail oneself of historical information for  attention calculation, while \cite{see2017get} provided a progressive version. \cite{nema2017diversity} introduced a series of \emph{diverse cell} structures to solve the duplicate words.

So far, few papers have considered about the structural or sementic issues at the language level in the field of summarization. 
\cite{filippova2008dependency} presented a novel unsupervised method that made use of a pruned dependency tree to acquire the sentence compression. Based on a Chinese short text summary dataset (LCSTS) and the \emph{attentional seq2seq model}, \cite{ma2017improving} proposed to enhance the semantic relevance by calculating the cos similarities of summaries and source texts.

\section{Conclusion}
In our paper, we construct a \emph{dual attentional seq2seq model} comprising source and extractive encoders to generate summaries. In addition, we put forward  the \emph{modified sentence ranking algorithm} to extract a specific number of high weighted sentences, for the purpose of strengthening the semantic representation of the extractive encoder. Furthermore, we introduce the \emph{pointer-generator} and \emph{coverage} mechanisms in our models so as to solve the problems of OOV and duplicate words. In the non-anonymized CNN/Daily Mail dataset, our results are close to the state-of-the-art ROUGE $\textbf{F}_1$ scores. Moreover, we get the highest abstractive ROUGE-2 $\textbf{F}_1$ scores, as well as obtain such summaries that have better readability and higher semantic accuracies. In our future work, we plan to unify the \emph{reinforcement learning} method with our abstractive models.

\section*{Acknowledgments}
We thank the anonymous reviewers for their insightful comments on this paper. 
This work was partially supported by National Natural Science Foundation of China (61572049 and 61333018). 
The correspondence author is Sujian Li.

%
%
%
\bibliographystyle{splncs04}
\bibliography{mybibliography}

\end{document}